\documentclass[10pt,twocolumn,letterpaper]{article}

\usepackage[pagenumbers]{wacv} 

\usepackage{graphicx}
\usepackage{amsmath}
\usepackage{amssymb}
\usepackage{booktabs}
\usepackage{colortbl}
\usepackage{textcomp}
\usepackage{xcolor}
 \usepackage{multirow}
\usepackage{subcaption}
 \usepackage{dblfloatfix}

 \usepackage[accsupp]{axessibility} 
 
\usepackage{bigstrut}
\usepackage{multirow}
\usepackage{color}
\usepackage{caption}
\usepackage{fancyhdr}
\usepackage{listings} 
\usepackage{changepage}
\usepackage{pgfplots}
\usepackage{balance}
\usepackage{enumitem}
\usepackage{breqn}
\usepackage[pagebackref=true,breaklinks=true,letterpaper=true,colorlinks,bookmarks=false]{hyperref}
\usepackage[capitalize]{cleveref}
\crefname{section}{Sec.}{Secs.}
\Crefname{section}{Section}{Sections}
\Crefname{table}{Table}{Tables}
\crefname{table}{Tab.}{Tabs.}

\def\wacvPaperID{359} 
\def\confName{WACV}
\def\confYear{2024}

\begin{document}

\title{Multispectral Imaging for Differential Face Morphing Attack Detection: A Preliminary Study}

\author{Raghavendra Ramachandra\textsuperscript{1} \quad  Sushma Venkatesh\textsuperscript{2} \quad   Naser Damer \textsuperscript{3} \quad   Narayan Vetrekar\textsuperscript{4} \quad  Rajendra Gad  \textsuperscript{4}\\
\textsuperscript{1} Norwegian University of Science and Technology (NTNU), Norway.
\textsuperscript{2}AiBA AS, Norway. \\
\textsuperscript{3} Fraunhofer Institute for Computer Graphics Research IGD, Darmstadt, Germany.
\textsuperscript{4}Goa University, India.\\
Email: raghavendra.ramachandra@ntnu.no
}


\maketitle

\begin{abstract}
Face morphing attack detection is emerging as an increasingly challenging problem owing to advancements in high-quality and realistic morphing attack generation. Reliable detection of morphing attacks is essential because these attacks are targeted for border control applications. This paper presents a multispectral framework for differential morphing-attack detection (D-MAD). The D-MAD methods are based on using two facial images that are captured from the ePassport (also called the reference image) and the trusted device (for example, Automatic Border Control (ABC) gates) to detect whether the face image presented in ePassport is morphed. The proposed multispectral D-MAD framework introduce a multispectral image captured as a trusted capture to acquire seven different spectral bands to detect morphing attacks. Extensive experiments were conducted on the newly created Multispectral Morphed Datasets (MSMD) with 143 unique data subjects that were captured using both visible and multispectral cameras in multiple sessions. The results indicate the superior performance of the proposed multispectral framework compared to visible images. 
\end{abstract}


\section{Introduction}
\label{sec:intro}

Face Recognition Systems (FRS) are widely deployed in numerous real-life access control applications. Face biometrics are extensively used in border control scenarios, resulting in more than one billion electric passports (or ePassports) \cite{ePassport} in which the face is used as the primary identifier. The exponential growth in the adaptation of ePassports and automatic Border Control (ABC) gates has also increased the risk of attacking these systems. Among the different types of attacks on ePassports, morphing attacks have emerged as potential attacks by deceiving both humans (at passport applications and border control) and ABC gates \cite{Venkatesh-MADSurvey-IEEETTS-2021, godage2022analyzing}. 

\begin{figure}[t!]
\begin{center}
\includegraphics[width=1.0\linewidth]{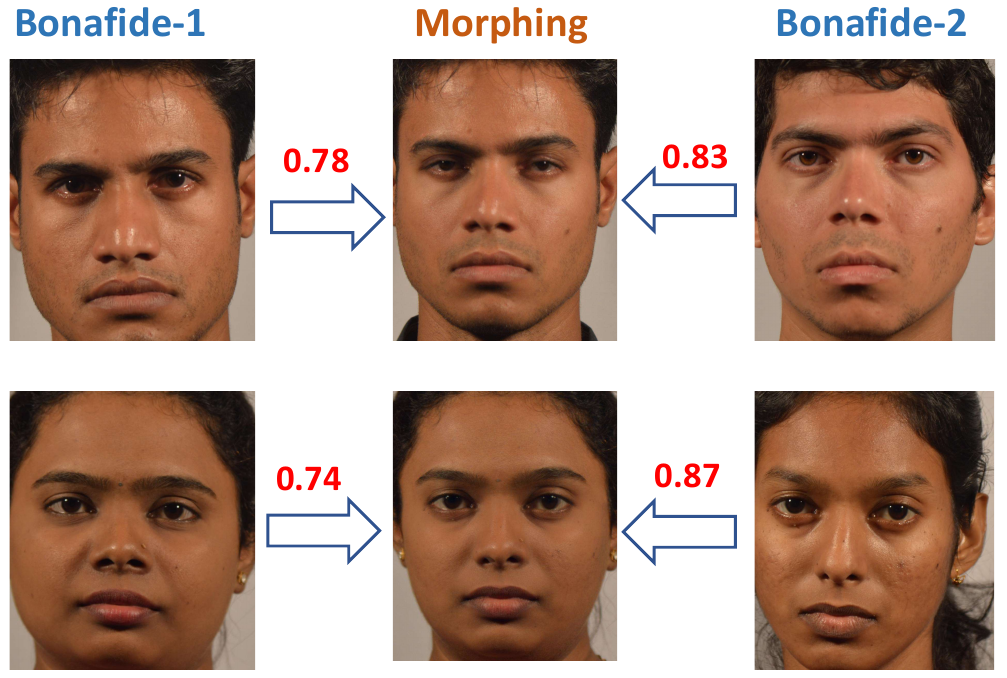}
\end{center}
   \caption{Example of face morphing images together with bona fide face images used for morphing. Comparison scores (higher is the better match) computed using Arcface FRS at FAR = $0.1\%$ is also illustrated.}
\label{fig:Intro}
\end{figure}

Morphing is defined as the process of seamlessly combining two or more face images from subjects such that the resulting morphing image indicates visual similarity to the source face images used for morphing. Extensive experiments reported in the literature \cite{godage2022analyzing, Makrushin-simulation-border-control-experiment-2020} show the vulnerability of FRS and the limitations of trained border guards in detecting morphed images. The attacker can use the morphing technique to generate the face images by blending his face image with the other data subject (also known as contributory data subjects) face image. The generated morphing image can then be used to obtain ID documents (e.g., ePassports) that can be used by all contributory data subjects. Morphing images can be generated using either landmark-based \cite{Ferrara-TextureBlendingAndShapeWarpingInFaceMorphing-IEEE-BIOSIG-2019} or deep learning based \cite{zhang-MIPGAN-TBIOM-2021} techniques. Furthermore, several open-source tools \cite{Venkatesh-MADSurvey-IEEETTS-2021} enable attackers to easily generate morphing attacks. Figure \ref{fig:Intro} shows the example of morphing images together with the bona fide images that are used to generate the morphing images. Therefore, reliable detection of the face morphing image is necessary to achieve reliable user verification using the FRS.

Morphing Attack Detection (MAD) has been extensively studied in the literature, resulting in two  types: single-image MAD (S-MAD)-based and differential-based MAD (D-MAD). S-MAD techniques are based on a single image to detect whether the presented face image is a bona fide or morph. The D-MAD methods are based on two images such that it uses the first image as the reference and then compares the given image with the reference image to detect the morphing attack. D-MAD techniques are well suited for the ABC scenario, where reference images can be taken from the passport and live captured images can be taken from the ABC gates. Various techniques are proposed for S-MAD, and D-MAD approaches \cite{Venkatesh-MADSurvey-IEEETTS-2021}. The existing S-MAD approaches include texture features \cite{Raghavendra-DetectingMorphedFace-BTAS-2016}, hybrid texture features \cite{Raghavendra-MADusingScaleSpaceFeature-ISBA-2019}, Residual noise \cite{Debiasi-PRNUVarianceMAD-BTAS-2018, Venkatesh-CANnetworkMAD-WACV-2020}, deep features \cite{medvedev2022mordeephy}, multimodal features \cite{raghavendra2022multimodality}, and attention models \cite{Aghdaie-waveletMAD-IJCB-2021}. The existing D-MAD approaches include de-morphing \cite{Ferrara-FaceDemorphing-IEEE-EUSIPCO-2018, Banerjee-conditional-GAN-IJCB-2021}, texture features \cite{siri-FusionMAD-Fusion-2021},  residual features \cite{ramachandra2022residual}, 3D information \cite{Singh-RobustMADatABCGate-SITIS-2019}, deep learning \cite{Borghi-DoubleSiamese-mdpi-2021, Soleymani-DMAD-MutualInfo-WACV2021, Peng-FD-GAN-IEEE-2019, soleymani-DMAD-Siamese-2021}, legacy \cite{Batskos21p_DMADLegacy_IET} and deep features \cite{Raghavendra-DNNMorphingDetection-CVPR-2017, Ortega-BorderControlMAD-IEEEAcess-2020, Scherhag-FaceMorphingAttacks-TIFS-2020}. For more details on the existing techniques for MAD readers, please refer to survey article \cite{Venkatesh-MADSurvey-IEEETTS-2021}. Furthermore, it is worth noting that D-MAD techniques have demonstrated better detection accuracy compared to S-MAD techniques.

Existing studies on S-MAD and D-MAD have been developed based on the visible face image, in which the face images are captured with conventional cameras. However, the evolving technology for developing ABC gates is equipped with multispectral cameras \cite{MSeGATE}. This motivated us to consider multi-spectral imaging as the source of image capture, especially in the D-MAD setup. Multispectral imaging captures multiple images in different spectral bands that are complementary to each other. Hence, it is our assertion that the fusion of complementary spectral bands improves the detection performance of D-MAD techniques. In this work, we present a novel framework for D-MAD in which the reference images are captured from visible imaging and the live image is captured using multispectral imaging. More particularly, we introduce the following critical questions:
\begin{itemize}[leftmargin=*,noitemsep, topsep=0pt,parsep=0pt,partopsep=0pt]
\item \textbf{Q1:} Which spectral band indicates the highest morphing detection accuracy? 
\item \textbf{Q2:} Does the individual multispectral imaging improves the morphing attack detection compared to the visible imaging alone?
\item \textbf{Q3:} Does the fusion of spectral bands improve the morphing attack detection compared to the visible imaging alone?
\end{itemize} 
While answering the above research questions, the following are the main contributions of this work:
\begin{itemize}[leftmargin=*,noitemsep, topsep=0pt,parsep=0pt,partopsep=0pt]
    \item To the best of our knowledge, this is the first study to address morphing attacks using multispectral imaging in the D-MAD framework. 
    \item The new multispectral dataset comprised 143 unique data subjects, of which 86 corresponded to male and 57 corresponded to female data subjects. Furthermore, a visible image dataset corresponding to 143 unique subjects was also collected in two different sessions. The database is publicly available for research purposes (\url{https://sites.google.com/view/narayanvetrekar/database/ spectral-face-gender}).
    \item Benchmarking the performance of two different D-MAD techniques based on deep features \cite{Scherhag-FaceMorphingAttacks-TIFS-2020} and hierarchical deep SLERP \cite{singh2022reliable}. 
\item Extensive experiments benchmarking morphing attack detection results with individual spectral bands and comparison with visible imaging.  
\end{itemize}
The remainder of this paper is organized as follows: Section \ref{sec:Msframework} discusses the proposed multispectral framework for the D-MAD, Section \ref{sec:MsDB} discusses the newly collected dataset with multi-spectral and visible samples, Section \ref{sec:exp} discusses the experimental results, and finally, Section \ref{sec:conc} draws conclusions and future work.

\section{Multi-spectral Differential  Morphing Attack Detection (D-MAD) Framework}
\label{sec:Msframework}

\begin{figure*}[htp]
\begin{center}
\includegraphics[width=1.0\linewidth]{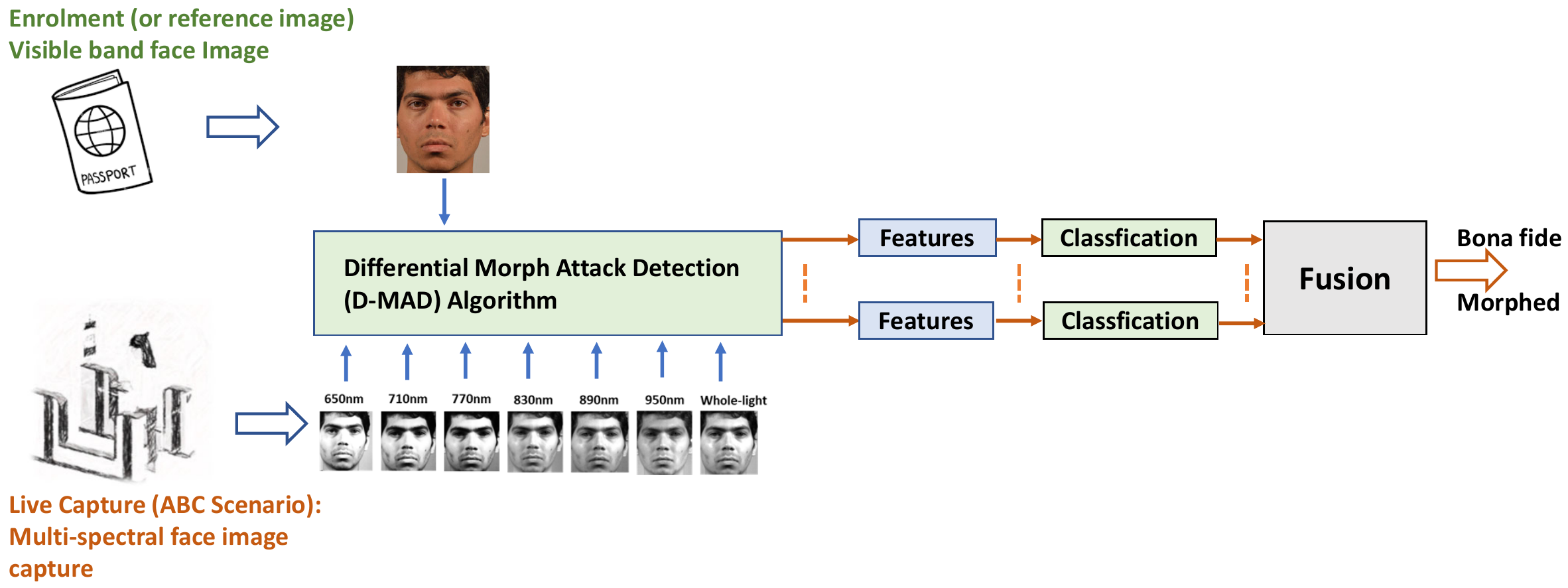}
\end{center}
   \caption{Block diagram of the proposed multispectral D-MAD framework}
\label{fig:Proposed}
\end{figure*}

 In this section, we present the proposed  multispectral D-MAD framework for reliable morphing attack detection.  We assert that the use of multispectral images will provide complementary information, as it can capture multiple images corresponding to different spectral bands. Furthermore, each of these spectral images reflects different levels of identity information because of the skin reflectance of different spectral bands. Thus, the use of multispectral imaging will enhance both the complementary and identity information of the data subject, which can result in the accurate detection of morphing attacks. 

Figure \ref{fig:Proposed} shows the block diagram of the proposed multispectral framework for Differential morphing attack detection (D-MAD). The D-MAD approach is based on two images: the first is the reference image and the second is the image captured from the trusted device. Therefore, the D-MAD scenario is highly applicable for  ABC gate border control applications where the reference image can be read from the epassport and the trusted  (or live capture) can be obtained from the ABC gate cameras. The enrolment image from passports is  from the visible spectrum. Because the applicant submits the passport application together with the passport photo, which is then re-digitized (scanned) to be stored in the passport. The trusted captured images in the proposed framework are from a multispectral camera that can render the images in the distinct spectral bands that complement each other. Each spectral image was used together with the reference image independently to extract the features that were classified to obtain the scores that were combined to make the final decision. The proposed framework can be structured into four functional units: (a) data capture, (b) D-MAD features, (c) classification scores and (d) fusion, as discussed below. 
\subsection{Data capture}
The enrolment images were captured using a DSLR camera (Nikon D320) in a photo studio set up with uniform lighting. The captured image was further processed to achieve compatibility with ICAO-9303 \cite{ICAO-9303-p9-2015} standards. Trusted device (or live) images were captured using a multispectral camera with a spatial resolution of 1.3Mpixels. In this work, we used six different spectral images captured at 650 nm, 710 nm, 770 nm, 830 nm, 890 nm, and 950 nm. Furthermore, we also captured a WholeLight (WL) image, which represents an image with no specific spectral filter. Let the reference image be $R$ and the multispectral trusted captured images be $T_{1}, \ldots, T_{N}$, where $N = 1,2, \ldots, 7$ corresponds to the spectral bands 650 nm, 710 nm, 770 nm, 830 nm, 890 nm, 950 nm and WL. More details on the data capture is discussed in the Section \ref{sec:MsDB}.
\subsection{D-MAD features}
The proposed framework can be used with existing D-MAD techniques because pairwise image processing is performed independently for individual spectral bands. Hence, in this study, we employed two D-MAD techniques to benchmark the performance of the proposed multispectral framework. The first D-MAD algorithm was based on deep features \cite{Scherhag-FaceMorphingAttacks-TIFS-2020} that are based on the Arcface features \cite{Deng-ArcFace-IEEE-CVPR-2019}. Given the reference image $R$ and the live image $T_{N}$, the deep features method extracts the face-related features using Arcface FRS on both $R$ and $T_{N}$ and provides difference features as the output. The second D-MAD method employed in this study is based on Hierarchical Deep Residual SLERP \cite{singh2022reliable}, which is based on six different pre-trained deep CNN networks. Furthermore, the feature differences between $R$ and $T_{N}$ were computed and combined using Spherical Linear Interpolation (SLERP) to output the final features.  In this study, we selected these two D-MAD techniques by considering their performance and generalizability in detecting morphing attacks. The deep features \cite{Scherhag-FaceMorphingAttacks-TIFS-2020} method also indicated good performance on the NIST benchmark, while Deep Residual SLERP \cite{singh2022reliable} indicated superior performance over other existing methods, especially in the unconstrained border control scenarios. 
\subsection{Classification scores}
The features computed from the individual D-MAD method corresponding to the individual spectral bands are then provided to the trained classifier to obtain the corresponding matching score. Because there are seven different spectral bands, we obtain seven different scores, $S_{1}, S_{2}, S_{3}, S_{4}, S_{5}, S_{6}, S_{7}$ corresponding to each feature. We used the same classifiers that were used with the existing D-MAD techniques employed in this study. 
\subsection{Fusion}
Finally, the proposed framework combines the scores obtained in the previous step using the sum rule. Thus, the fused scores $F = \sum_{k=1}^{7} S_{k}$ is compared with the pre-set threshold to make the final decision on the morphing detection.  

\section{Multispectral Morphing Dataset (MSMD)}
\label{sec:MsDB}

\begin{figure*}[htp]
\begin{center}
\includegraphics[width=1.0\linewidth]{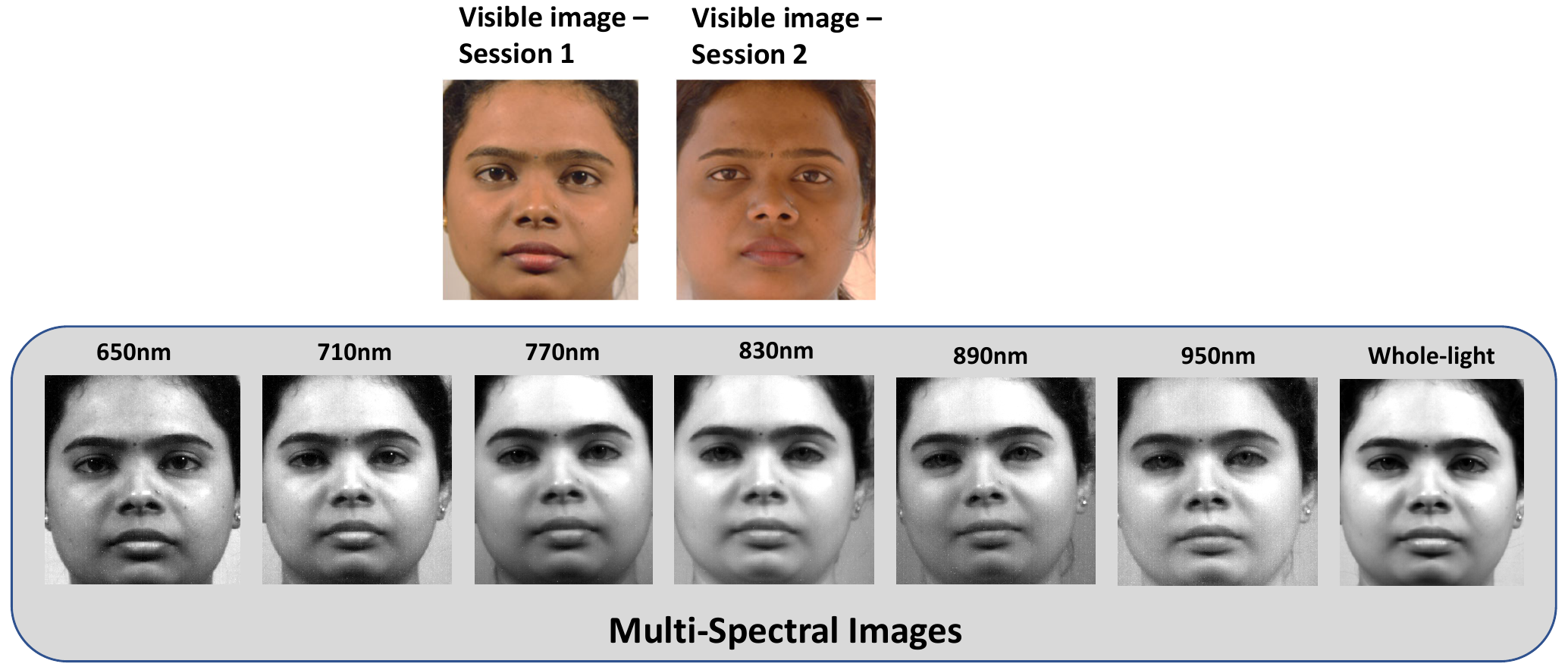}
\end{center}
   \caption{Example of visible (from session-1 and session-2) and multispectral images from MSMD dataset}
\label{fig:db}
\end{figure*}

This section presents a newly collected Multispectral Morphing dataset (MSMD). The MSMD dataset comprised 143 subjects (86 males and 57 females) with varying age groups from  22 to 72 years. The MSMD dataset was collected in multiple sessions to capture visible and multispectral images from 143 subjects. 
\subsection {Visible data collection:} The visible images are collected using DSLR camera (Model: Nikon D320) under constrained conditions. Because visible images represent passport images, special attention is required to achieve high-quality face image capture. Visible images were captured using a photo booth with uniform lighting, neutral pose, and expression. Visible data were captured in two sessions with a time gap of 30-45 days. Sesison-1 was collected under highly constrained conditions, while session-2 was collected under similar conditions. Session-1 is used as the enrolment samples and session-2 samples are used as the trusted capture in the experiments. In session-1, 13 images were captured and in session-2, 30 image samples were  captured. Thus, the visible dataset comprised $143$ subjects $ \times$ $13$ samples = $1859$ samples in session-1 and $143$ subjects $ \times$ $30$ samples = $4290$ samples in session-2. Figure \ref{fig:db} shows the example visible images from session-1 and session-2 from the MSMD dataset. 
\subsection{Multispectral data collection:} The multispectral data were collected using a custom device built using a CMOS camera (Model: BCi5-U-M-40-LP) with 1.3 Mega pixels spatial resolution.  In this study, we selected six different spectral bands, 650 nm, 710 nm, 770 nm, 830 nm, 890 nm, and 950 nm, together with the WholeLight (WL) image. We selected these six spectral bands to cover both visible (VIS) and near-infrared (NIR) wavelengths. Furthermore, the availability of the WholeLight image, in which the image is captured without any filters, represents the standard image captured with the VIS-NIR-sensitive camera. The dataset is collected in the indoor setting with two illumination units of a quartz tungsten halogen (QTH) light source. We have collected 11 samples for each data subject that will result in $143$ subjects $\times$ $7$ spectral bands $\times$ $11$ = 11,011 samples. Figure \ref{fig:db} shows the example multispectral images from the MSMD dataset.

\subsection{Morphing images:} The morphing images were generated using the visible images collected in session-1 of the visible image dataset. We selected one image per data subject and employed two morphing generation methods to create morphing attack samples. The first morphing generation is based on using landmarks \cite{Ferrara-TextureBlendingAndShapeWarpingInFaceMorphing-IEEE-BIOSIG-2019}  in which pixel-level information from the contributing data subjects is employed by wrapping and blending to generate high-quality morph images. The second morphing generation is based on a generative method called MIPGAN-2 \cite{zhang-MIPGAN-TBIOM-2021}, which uses GAN-2 as the backbone. The motivations behind the selection of these methods include (a) high-quality morphing image generation, (b) indicating high vulnerability with FRS  \cite{zhang-MIPGAN-TBIOM-2021} and (c) challenges to be detected using MAD techniques. \cite{zhang-MIPGAN-TBIOM-2021}. 

\begin{figure}[htp]
\begin{center}
\includegraphics[width=1.0\linewidth]{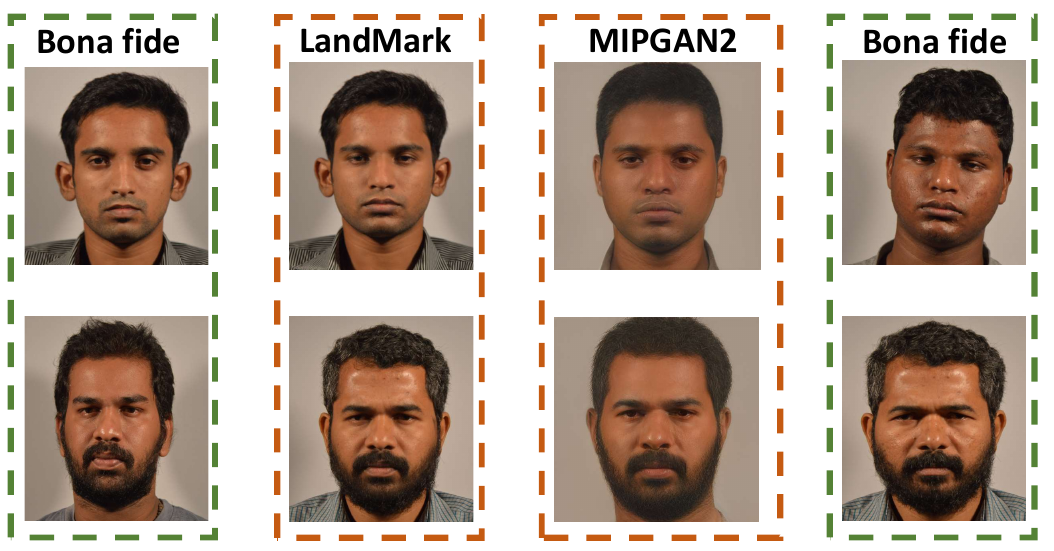}
\end{center}
   \caption{Example of face morphing images generated using Landmark and MIPGAN2 based morphing techniques.}
\label{fig:MorphImages}
\end{figure}
 To ensure a disjoint-morphing dataset, the dataset of 143 unique subjects was divided into two disjoint sets. The training set consisted of 78 data subjects and the testing set consisted of 65 datasets. Morphing is performed internally on the training and testing sets to achieve a complete disjoint set. In total, we had 1400 morphing images, of which the training set consisted of 928 images and the testing set consisted of 472 morphing images. Figure \ref{fig:MorphImages} shows an example of morphing the images from the MSMD dataset. Table \ref{tab:statsMSMD} lists the statistics of the MSMD dataset, which is available to the semi-public for research purposes. 

\begin{table}[htp]
  \centering
  \caption{Statistics of MSMD dataset summarising the data partition for training and testing set}
  \resizebox{0.89\linewidth}{!}{
    \begin{tabular}{|p{9.215em}|c|c|}
    \hline
    Data Type & \multicolumn{1}{p{3.215em}|}{Training Set} & \multicolumn{1}{p{4.215em}|}{Testing Set} \bigstrut\\
    \hline
    Visible Images & 1859  & 4290 \bigstrut\\
    \hline
    Multispectral images & 5698  & 5313 \bigstrut\\
    \hline
    Morphing Images & 928 $\times$ 2  & 472 $\times$ 2 \bigstrut\\
    \hline
    \end{tabular}%
    }
  \label{tab:statsMSMD}%
\end{table}%
\begin{figure*}[htp]
\minipage{0.47\textwidth}
  \includegraphics[width=\linewidth]{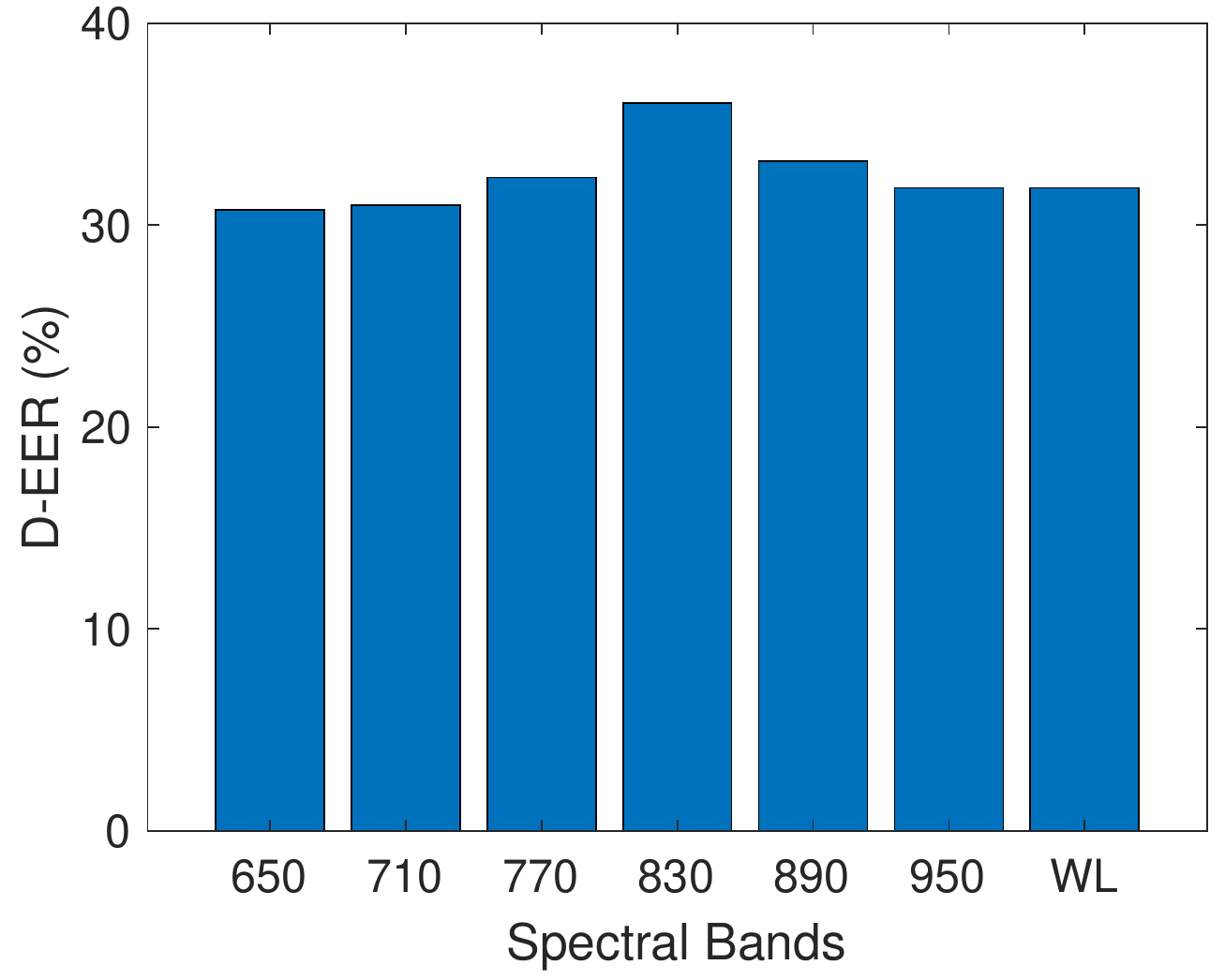}
  \caption{D-EER(\%) performance of Deep features \cite{Scherhag-FaceMorphingAttacks-TIFS-2020} on individual spectral bands: Landmark based morphing \cite{Ferrara-TextureBlendingAndShapeWarpingInFaceMorphing-IEEE-BIOSIG-2019}}\label{fig:Arc}
\endminipage
\hfill
\minipage{0.465\textwidth}
  \includegraphics[width=\linewidth]{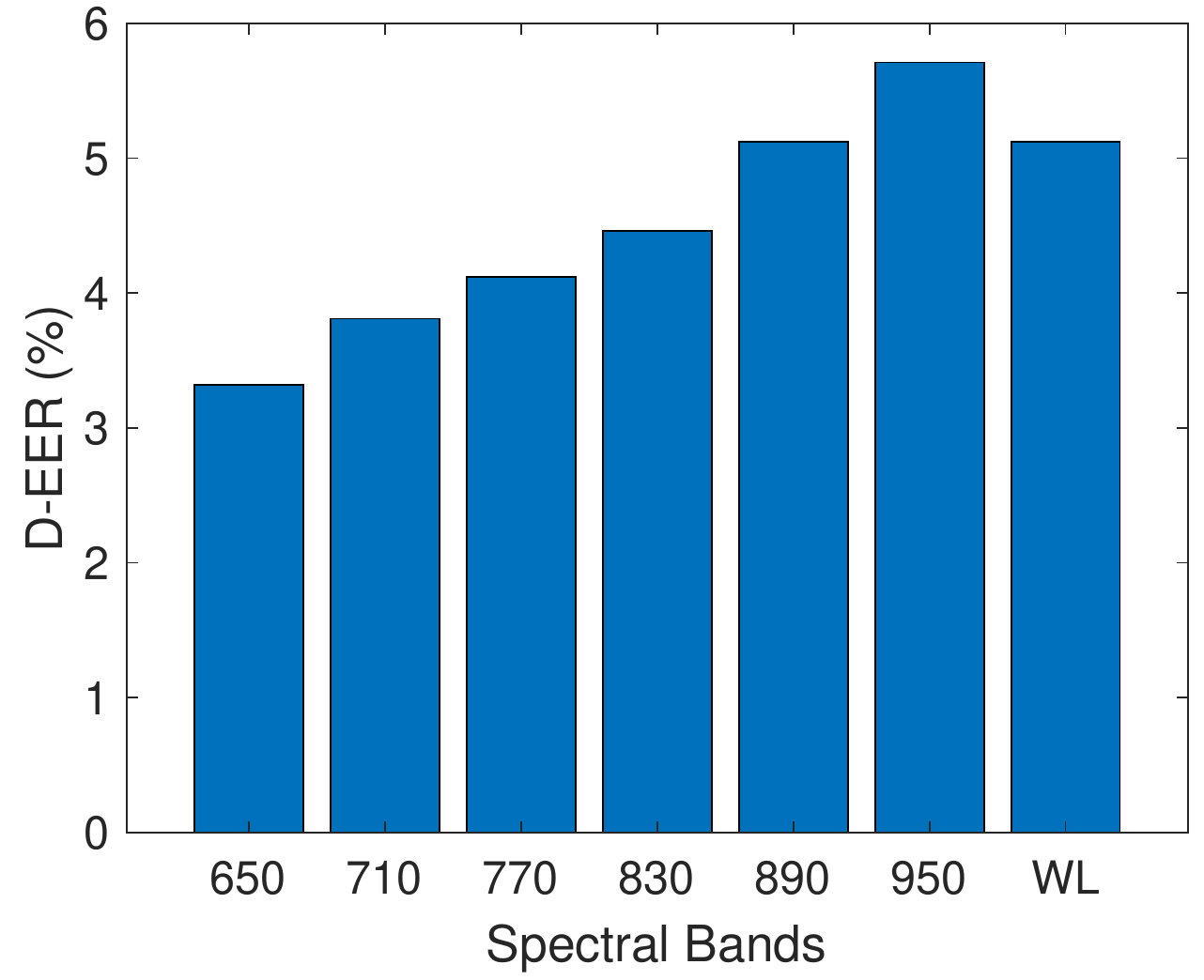}
  \caption{D-EER(\%) performance of Hierarchical deep Residual SLERP \cite{singh2022reliable} on individual spectral bands: Landmark based morphing \cite{Ferrara-TextureBlendingAndShapeWarpingInFaceMorphing-IEEE-BIOSIG-2019}} \label{fig:slerp}
\endminipage
\label{fig:DET}%
\end{figure*}

\section{Experiments and Results}
\label{sec:exp}
In this section, we present quantitative results of the proposed multispectral D-MAD framework. The quantitative performance of the D-MAD techniques is presented using the ISO/IEC 30107 metrics \cite{ISO-IEC-30107-3-PAD-metrics-170227} namely the ``Attack Presentation Classification Error Rate (APCER ($\%$)), which defines the proportion of attack images (face morphing images) incorrectly classified as bona fide images and the Bona fide Presentation Classification Error Rate (BPCER ($\%$)) in which bona fide images incorrectly classified as attack images are counted \cite{ISO-IEC-30107-3-PAD-metrics-170227}  along with the Detection Equal Error Rate (D-EER ($\%$))" \cite{zhang-MIPGAN-TBIOM-2021}. 
\subsection{Experimental protocols}
We independently performed two experiments on the visible and multispectral data to benchmark the comparative performance of the D-MAD techniques employed in this study. 
\textbf{Visible D-MAD experiments:} In this experiment, session-1 visible images represent the bona fide samples, and the corresponding morphed images as the morphing samples are used as the enrolment (or reference) image. The trusted capture device image is represented by the visible data captured in session-2. 
\textbf{Multispectral D-MAD experiments:} In this experiment, the reference images corresponding to  bona fide and morphing samples were taken from the visible images collected in session-1. Trusted capture images were obtained from the multispectral data. 

\begin{table*}[htp]
  \centering
\caption{Quantitative detection performance of the D-MAD algorithms on individual spectral bands}
  \resizebox{0.95\linewidth}{!}{
\begin{tabular}{|c|c|c|c|cc|}
\hline
\rowcolor[HTML]{ECF4FF} 
\cellcolor[HTML]{ECF4FF}                                                     & \cellcolor[HTML]{ECF4FF}                                          & \cellcolor[HTML]{ECF4FF}                                          & \cellcolor[HTML]{ECF4FF}                                      & \multicolumn{2}{c|}{\cellcolor[HTML]{ECF4FF}\textbf{BPCER @APCER =}}      \\ \cline{5-6} 
\rowcolor[HTML]{ECF4FF} 
\multirow{-2}{*}{\cellcolor[HTML]{ECF4FF}\textbf{Morphing Type}}             & \multirow{-2}{*}{\cellcolor[HTML]{ECF4FF}\textbf{MAD Algorithms}} & \multirow{-2}{*}{\cellcolor[HTML]{ECF4FF}\textbf{Spectral Bands}} & \multirow{-2}{*}{\cellcolor[HTML]{ECF4FF}\textbf{D-EER (\%)}} & \multicolumn{1}{c|}{\cellcolor[HTML]{ECF4FF}\textbf{5\%}} & \textbf{10\%} \\ \hline
\cellcolor[HTML]{FFCCC9}                                                     &                                                                   & 650                                                               & 30.75                                                         & \multicolumn{1}{c|}{85.82}                                & 69.33         \\ \cline{3-6} 
\cellcolor[HTML]{FFCCC9}                                                     &                                                                   & 710                                                               & 30.98                                                         & \multicolumn{1}{c|}{87.18}                                & 71.19         \\ \cline{3-6} 
\cellcolor[HTML]{FFCCC9}                                                     &                                                                   & 770                                                               & 32.35                                                         & \multicolumn{1}{c|}{84.68}                                & 70.14         \\ \cline{3-6} 
\cellcolor[HTML]{FFCCC9}                                                     &                                                                   & 830                                                               & 36.14                                                         & \multicolumn{1}{c|}{89.66}                                & 84.3          \\ \cline{3-6} 
\cellcolor[HTML]{FFCCC9}                                                     &                                                                   & 890                                                               & 33.13                                                         & \multicolumn{1}{c|}{84.33}                                & 72.16         \\ \cline{3-6} 
\cellcolor[HTML]{FFCCC9}                                                     &                                                                   & 950                                                               & 31.84                                                         & \multicolumn{1}{c|}{85.67}                                & 69.85         \\ \cline{3-6} 
\cellcolor[HTML]{FFCCC9}                                                     & \multirow{-7}{*}{\textbf{Deep Features \cite{Scherhag-FaceMorphingAttacks-TIFS-2020}}}                          & WL                                                                & 29.61                                                         & \multicolumn{1}{c|}{84.71}                                & 68.81         \\ \cline{2-6} 
\cellcolor[HTML]{FFCCC9}                                                     &                                                                   & 650                                                               & 3.32                                                          & \multicolumn{1}{c|}{1.69}                                 & 0.49          \\ \cline{3-6} 
\cellcolor[HTML]{FFCCC9}                                                     &                                                                   & 710                                                               & 3.81                                                          & \multicolumn{1}{c|}{2.98}                                 & 1.22          \\ \cline{3-6} 
\cellcolor[HTML]{FFCCC9}                                                     &                                                                   & 770                                                               & 4.22                                                          & \multicolumn{1}{c|}{2.71}                                 & 1.66          \\ \cline{3-6} 
\cellcolor[HTML]{FFCCC9}                                                     &                                                                   & 830                                                               & 4.46                                                          & \multicolumn{1}{c|}{3.74}                                 & 1.14          \\ \cline{3-6} 
\cellcolor[HTML]{FFCCC9}                                                     &                                                                   & 890                                                               & 4.99                                                          & \multicolumn{1}{c|}{4.98}                                 & 2.63          \\ \cline{3-6} 
\cellcolor[HTML]{FFCCC9}                                                     &                                                                   & 950                                                               & 5.5                                                           & \multicolumn{1}{c|}{5.74}                                 & 3.34          \\ \cline{3-6} 
\multirow{-14}{*}{\cellcolor[HTML]{FFCCC9}\textbf{Land Mark based Morphing \cite{Ferrara-TextureBlendingAndShapeWarpingInFaceMorphing-IEEE-BIOSIG-2019}}} & \multirow{-7}{*}{\textbf{Hierarchical Deep Residual SLERP \cite{singh2022reliable}}}       & WL                                                                & 5.12                                                          & \multicolumn{1}{c|}{5.2}                                  & 2.63          \\ \hline \hline
\cellcolor[HTML]{CBCEFB}                                                     &                                                                   & 650                                                               & 8.77                                                          & \multicolumn{1}{c|}{18.56}                                & 6.72          \\ \cline{3-6} 
\cellcolor[HTML]{CBCEFB}                                                     &                                                                   & 710                                                               & 9.41                                                          & \multicolumn{1}{c|}{23.97}                                & 8.55          \\ \cline{3-6} 
\cellcolor[HTML]{CBCEFB}                                                     &                                                                   & 770                                                               & 11.17                                                         & \multicolumn{1}{c|}{29.81}                                & 13.19         \\ \cline{3-6} 
\cellcolor[HTML]{CBCEFB}                                                     &                                                                   & 830                                                               & 11.73                                                         & \multicolumn{1}{c|}{27.60}                                & 13.98         \\ \cline{3-6} 
\cellcolor[HTML]{CBCEFB}                                                     &                                                                   & 890                                                               & 12.68                                                         & \multicolumn{1}{c|}{28.93}                                & 17.25         \\ \cline{3-6} 
\cellcolor[HTML]{CBCEFB}                                                     &                                                                   & 950                                                               & 9.89                                                          & \multicolumn{1}{c|}{25.75}                                & 9.82          \\ \cline{3-6} 
\cellcolor[HTML]{CBCEFB}                                                     & \multirow{-7}{*}{\textbf{Deep Features \cite{Scherhag-FaceMorphingAttacks-TIFS-2020}}}                          & WL                                                                & 10.63                                                         & \multicolumn{1}{c|}{24.52}                                & 11.59         \\ \cline{2-6} 
\cellcolor[HTML]{CBCEFB}                                                     &                                                                   & 650                                                               & \textbf{0}                                                    & \multicolumn{1}{c|}{\textbf{0}}                           & \textbf{0}    \\ \cline{3-6} 
\cellcolor[HTML]{CBCEFB}                                                     &                                                                   & 710                                                               & \textbf{0}                                                    & \multicolumn{1}{c|}{\textbf{0}}                           & \textbf{0}    \\ \cline{3-6} 
\cellcolor[HTML]{CBCEFB}                                                     &                                                                   & 770                                                               & \textbf{0}                                                    & \multicolumn{1}{c|}{\textbf{0}}                           & \textbf{0}    \\ \cline{3-6} 
\cellcolor[HTML]{CBCEFB}                                                     &                                                                   & 830                                                               & \textbf{0}                                                    & \multicolumn{1}{c|}{\textbf{0}}                           & \textbf{0}    \\ \cline{3-6} 
\cellcolor[HTML]{CBCEFB}                                                     &                                                                   & 890                                                               & \textbf{0}                                                    & \multicolumn{1}{c|}{\textbf{0}}                           & \textbf{0}    \\ \cline{3-6} 
\cellcolor[HTML]{CBCEFB}                                                     &                                                                   & 950                                                               & \textbf{0}                                                    & \multicolumn{1}{c|}{\textbf{0}}                           & \textbf{0}    \\ \cline{3-6} 
\multirow{-14}{*}{\cellcolor[HTML]{CBCEFB}\textbf{MIPGAN-2 based Morphing \cite{zhang-MIPGAN-TBIOM-2021}}}  & \multirow{-7}{*}{\textbf{Hierarchical Deep Residual SLERP \cite{singh2022reliable}}}       & WL                                                                & \textbf{0}                                                    & \multicolumn{1}{c|}{\textbf{0}}                           & \textbf{0}    \\ \hline
\end{tabular}
}
\label{tab:indiSpectral}%
\end{table*}

\begin{table*}[htp]
 \centering
  \caption{Quantitative performance of D-MAD techniques on visible and multispectral data}
    \resizebox{0.89\linewidth}{!}{
\begin{tabular}{|l|l|l|l|ll|}
\hline
\rowcolor[HTML]{ECF4FF} 
\cellcolor[HTML]{ECF4FF}                                           & \cellcolor[HTML]{ECF4FF}                                     & \cellcolor[HTML]{ECF4FF}                                          & \cellcolor[HTML]{ECF4FF}                                      & \multicolumn{2}{l|}{\cellcolor[HTML]{ECF4FF}\textbf{BPCER @APCER =}}      \\ \cline{5-6} 
\rowcolor[HTML]{ECF4FF} 
\multirow{-2}{*}{\cellcolor[HTML]{ECF4FF}\textbf{Morphing Type}}   & \multirow{-2}{*}{\cellcolor[HTML]{ECF4FF}\textbf{Data Type}} & \multirow{-2}{*}{\cellcolor[HTML]{ECF4FF}\textbf{MAD Algorithms}} & \multirow{-2}{*}{\cellcolor[HTML]{ECF4FF}\textbf{D-EER (\%)}} & \multicolumn{1}{l|}{\cellcolor[HTML]{ECF4FF}\textbf{5\%}} & \textbf{10\%} \\ \hline
\cellcolor[HTML]{EFEFEF}                                           &                                                              & Deep Features         \cite{Scherhag-FaceMorphingAttacks-TIFS-2020}                                            & 32.69                                                         & \multicolumn{1}{l|}{83.61}                                & 69.58         \\ \cline{3-6} 
\cellcolor[HTML]{EFEFEF}                                           & \multirow{-2}{*}{Visible}                                    & Hierarchical Deep Residual SLERP  \cite{singh2022reliable}                                & 5.15                                                          & \multicolumn{1}{l|}{5.19}                                 & 2.17          \\ \cline{2-6} 
\cellcolor[HTML]{EFEFEF}                                           &                                                              & Deep Features  \cite{Scherhag-FaceMorphingAttacks-TIFS-2020}                                                   & 21.64                                                         & \multicolumn{1}{l|}{48.51}                                & 34.37         \\ \cline{3-6} 
\multirow{-4}{*}{\cellcolor[HTML]{EFEFEF}Land Mark based Morphing \cite{Ferrara-TextureBlendingAndShapeWarpingInFaceMorphing-IEEE-BIOSIG-2019}} & \multirow{-2}{*}{Multispectral}                              & Hierarchical Deep Residual SLERP \cite{singh2022reliable}                                 & 2.51                                                          & \multicolumn{1}{l|}{1.51}                                 & 0.34          \\ \hline \hline
\cellcolor[HTML]{FFCE93}                                           &                                                              & Deep Features \cite{Scherhag-FaceMorphingAttacks-TIFS-2020}                                                    & 17.73                                                         & \multicolumn{1}{l|}{36.69}                                & 26.32         \\ \cline{3-6} 
\cellcolor[HTML]{FFCE93}                                           & \multirow{-2}{*}{Visible}                                    & Hierarchical Deep Residual SLERP  \cite{singh2022reliable}                                & 3.91                                                          & \multicolumn{1}{l|}{3.01}                                 & 1.33          \\ \cline{2-6} 
\cellcolor[HTML]{FFCE93}                                           &                                                              & Deep Features  \cite{Scherhag-FaceMorphingAttacks-TIFS-2020}                                                   & 3.88                                                          & \multicolumn{1}{l|}{2.92}                                 & 0.87          \\ \cline{3-6} 
\multirow{-4}{*}{\cellcolor[HTML]{FFCE93}MIPGAN-2 based Morphing \cite{zhang-MIPGAN-TBIOM-2021}}  & \multirow{-2}{*}{Multispectral}                              & Hierarchical Deep Residual SLERP  \cite{singh2022reliable}                                & \textbf{0}                                                             & \multicolumn{1}{l|}{\textbf{0}}                                    & \textbf{0}             \\ \hline
\end{tabular}
}
\label{tab:Results}%
\end{table*}


\begin{figure*}[htp]
\minipage{0.47\textwidth}
  \includegraphics[width=\linewidth]{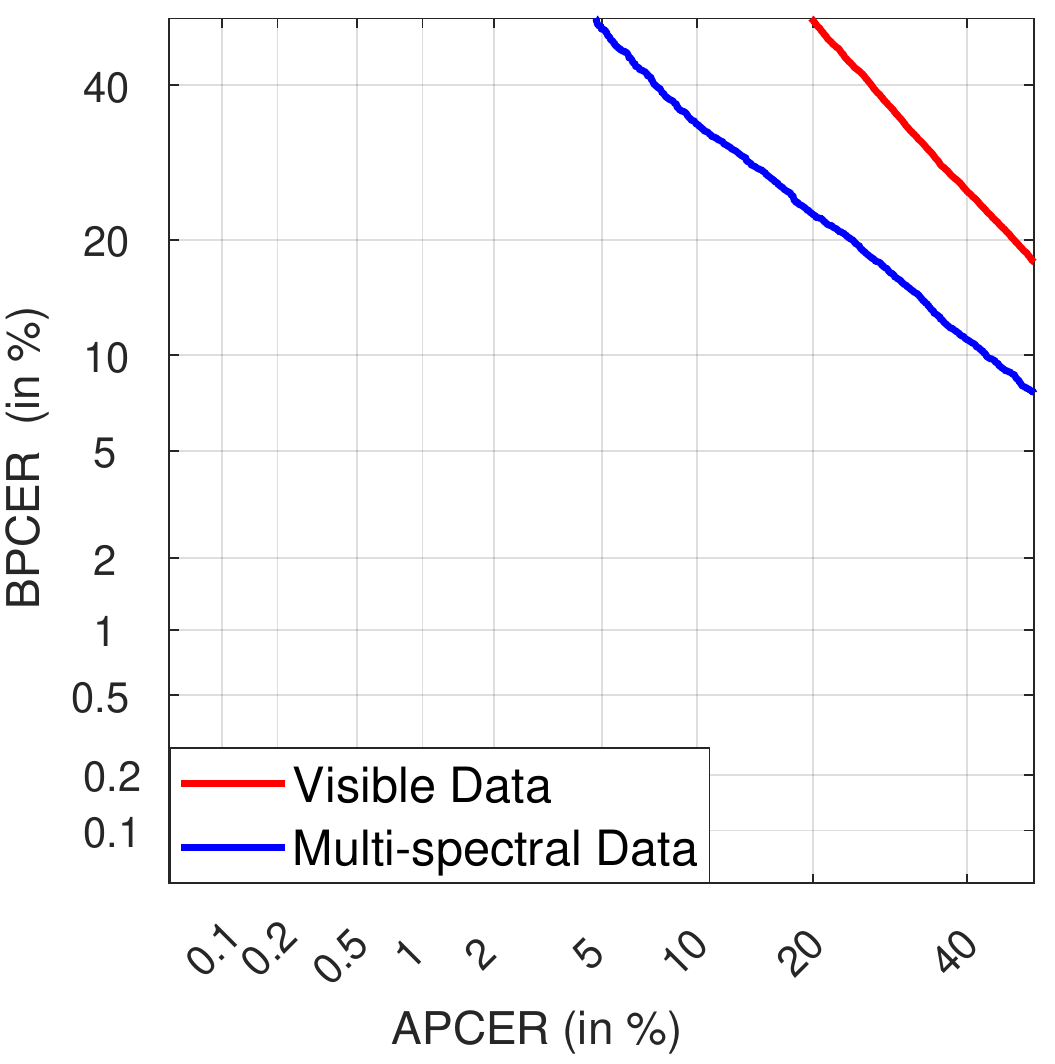}
  \caption{DET performance of Deep features \cite{Scherhag-FaceMorphingAttacks-TIFS-2020} on visible and multispectral data: Landmark based morphing \cite{Ferrara-TextureBlendingAndShapeWarpingInFaceMorphing-IEEE-BIOSIG-2019} }\label{fig:DET_ARC}
\endminipage
\hfill
\minipage{0.465\textwidth}
  \includegraphics[width=\linewidth]{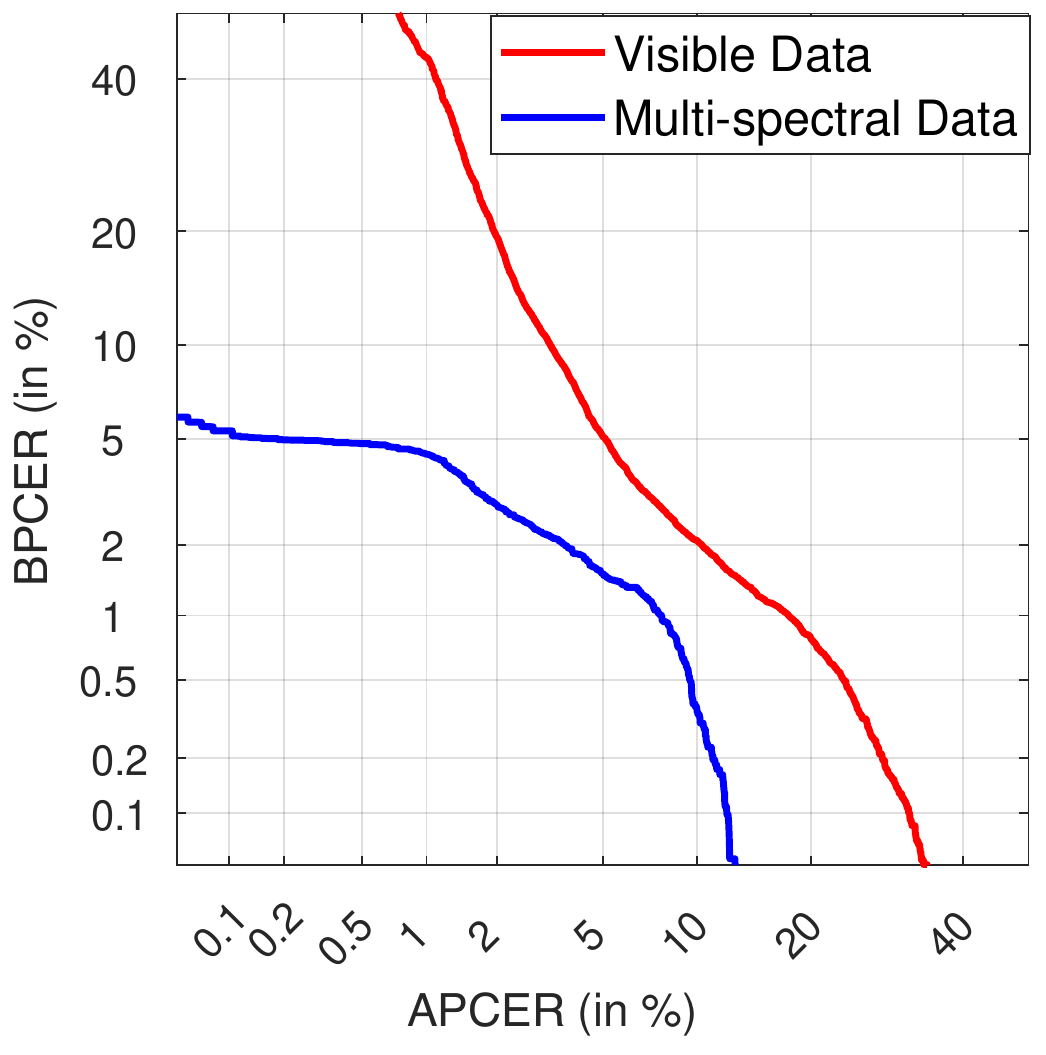}
  \caption{DET performance of Hierarchical Deep Residual SLERP \cite{singh2022reliable} on visible and multispectral data: Landmark based morphing \cite{Ferrara-TextureBlendingAndShapeWarpingInFaceMorphing-IEEE-BIOSIG-2019}}\label{fig:DET_SLERP}
\endminipage
\end{figure*}
\subsection{Results and Discussion}
First, we present the results of the D-MAD algorithms for the individual spectral bands of the multispectral data. Table \ref{tab:indiSpectral} shows the quantitative detection performance of the D-MAD techniques on two different types of morphing generation methods.  Figure \ref{fig:Arc} and \ref{fig:slerp} show bar charts of the D-EER (\%) corresponding to the individual spectral bands.  For simplicity, we included bar charts for the landmark-based morphing generation method.   Based on the obtained results, the following observations were made: 

\begin{itemize}
   \item The detection performance of the D-MAD techniques varies across different spectral bands. The variation in the detection performance was noted for both D-MAD techniques evaluated in this work. 
   \item The detection performance is also influenced by the morphing generation technique used to generate morphing. The experimental results indicate that MIPGAN-based morphing images are easier to detect than landmark-based morphing methods.  This observation is consistent with both visible and multispectral imaging. 
\item Among different spectral bands, the visible range bands (650nm and 710nm) indicates the lowest D-EER (\%) with both D-MAD techniques. The improved performance can be attributed to the fact that both enrolment and trusted images are from the visible spectrum, and facial images possess rich texture information that can aid the detection process. This observation is consistent with both visible and multispectral imaging. 
\item The detection performance of D-MAD in the near-infrared spectral bands (830 nm, 890 nm, and 950 nm) indicates a slightly higher D-EER (\%) compared to the visible spectral bands with landmark-based morphing. However, the variation in D-EER (\%) was not consistent across the two D-MAD techniques employed in this study. The deep features method \cite{Scherhag-FaceMorphingAttacks-TIFS-2020}, which is based on facial features, exhibits less variation in detection performance, especially in the NIR bands. However, the Hierarchical Deep Residual SLERP \cite{singh2022reliable}, which is based on texture features, indicates smaller variation in the detection performance, especially in NIR spectral bands, compared to the deep features method \cite{Scherhag-FaceMorphingAttacks-TIFS-2020}. The possible variation in the Hierarchical Deep Residual SLERP \cite{singh2022reliable} can be attributed to the lack of texture information in the NIR spectral bands. Thus, the detection performance of D-MAD in the NIR spectrum depends on the type of D-MAD features used for morphing detection. The use of identity features (as in \cite{Scherhag-FaceMorphingAttacks-TIFS-2020}) can indicate consistent performance across all spectral bands. However, the Hierarchical Deep Residual SLERP \cite{singh2022reliable} D-MAD has indicated the better performance across all spectral bands compared to the deep features method \cite{Scherhag-FaceMorphingAttacks-TIFS-2020}.   
\item The use of wholeLight (WL) that is captured without any spectral filtering indicates the varied detection performance with respect to D-MAD techniques.    The detection error with the deep features method \cite{Scherhag-FaceMorphingAttacks-TIFS-2020} on WL images indicates a performance similar to that of both the VIS and NIR spectral bands. Because the deep features method \cite{Scherhag-FaceMorphingAttacks-TIFS-2020} is based on facial features, it is less sensitive to different spectral bands, which may be due to the backbone model that is trained only using VIS images. However, the performance of Hierarchical Deep Residual SLERP \cite{singh2022reliable} on WL images shows degraded performance compared to the VIS spectral bands.  
\end{itemize}

Table \ref{tab:Results} lists the performance of the  D-MAD algorithms on the proposed multispectral framework and visible images. Both multispectral and visible data are collected using the same data subjects, and this can provide insights into the utility of multispectral images for reliable morphing attack detection. Furthermore, the performance of the multispectral framework was presented by fusing the results of all individual spectral bands. Figure \ref{fig:DET_ARC} and \ref{fig:DET_SLERP} show the D-MAD technique  DET curves for both the visible and multispectral data. The following are critical observations based on the results.

\begin{itemize}
\item In general, the performance of the D-MAD algorithms indicates a superior detection accuracy on the multispectral data compared to the visible data. This can be attributed to the fusion of complementary spectral bands, which can contribute to a higher detection accuracy. Improved detection performance can be noticed with both D-MAD techniques on multispectral data. 
\item The D-MAD method based on the Hierarchical Deep Residual SLERP \cite{singh2022reliable} indicates the superior performance on visible and multispectral data. 
\item Landmark-based morphing is more challenging to detect than MIPGAN-based morphing techniques. However, the proposed multispectral approach indicated higher detection accuracy for both morphing types. These results justify the efficacy of the proposed multispectral approach for face morphing detection. 

\end{itemize}

\subsection{Discussion}

Based on the observations from the above experiments and obtained results, the research questions formulated in Section \ref{sec:intro} are answered below.
\begin{itemize}[leftmargin=*,noitemsep, topsep=0pt,parsep=0pt,partopsep=0pt]

\item {\textbf{Q1}. Which spectral band indicates the highest morphing detection accuracy?} 
        \begin{itemize}[leftmargin=*,noitemsep, topsep=0pt,parsep=0pt,partopsep=0pt]
            \item Based on the experimental results reported in Table \ref{tab:indiSpectral},  the visible spectral bands indicate better detection accuracy on the textures-based D-MAD \cite{singh2022reliable}. However, the performance across the spectral bands is comparable when D-MAD with facial features \cite{Scherhag-FaceMorphingAttacks-TIFS-2020} is used. The performance of the D-MAD techniques varies with the type of morphing generation technique. The multispectral D-MAD approach when morphing generation is MIPGAN, indicates good  detection performance across all spectral bands. 
        \end{itemize}

    \item {\textbf{Q2}. Does the individual multispectral imaging improves the morphing attack detection compared to the visible imaging alone?}
    \begin{itemize}[leftmargin=*,noitemsep, topsep=0pt,parsep=0pt,partopsep=0pt]
        \item Based on the quantitative performance reported in Tables \ref{tab:Results} and \ref{tab:indiSpectral}, the individual spectral bands indicate slightly better performance compared to the visible images alone. Improved performance of the individual spectral bands can be observed with both D-MAD techniques.  However, with MIPGAN-based morphing, individual spectral bands indicate higher detection accuracy than the visible band alone with D-MAD approaches. 
    \end{itemize}

 \item {\textbf{Q3:} Does the fusion of spectral bands improves the morphing attack detection compared to the visible imaging alone?} 
        \begin{itemize}[leftmargin=*,noitemsep, topsep=0pt,parsep=0pt,partopsep=0pt]
            \item Based on the quantitative performance reported in Table \ref{tab:Results}, multispectral images indicate the highest detection accuracy compared to visible images. The improved performance is confirmed with two different D-MAD algorithms that are based on facial features \cite{Scherhag-FaceMorphingAttacks-TIFS-2020} and textures features \cite{singh2022reliable}. Hence, irrespective of the type of features used by the D-MAD techniques, the proposed multispectral framework indicated  improved morphing-detection accuracy with different types of morphing-generation methods.   
        \end{itemize}   
    
    \end{itemize}

\section{Conclusion}
\label{sec:conc}

In this paper, we presented a multispectral framework for differential morphing attack detection. The D-MAD framework utilizes two images. The first image is taken from ePassport (a.k.a reference image), and the second image is taken from the trusted capture (e.g., ABC gates) to detect the morphing attack on ePassport. The proposed framework captures seven different spectral bands as the trusted capture, which is further used with the reference image from the visible spectrum to reliably detect morphing attacks on ePassport. In this study, two different D-MAD techniques based on facial and texture features are employed within the proposed multispectral framework to compare the morphing detection performance of the multispectral approach with visible images. Extensive experiments were conducted on a newly built dataset with 143 unique data subjects. The obtained results demonstrate the superior performance of the proposed multispectral D-MAD compared to visible images on two different types of morphing generation techniques . Future work will extend the present work (a) with different print and scan sources for morphing image generation and (2) with different light sources for multispectral image capture.

{\small
\bibliographystyle{ieee_fullname}
\bibliography{Face_Morph_references}
}

\end{document}